\documentclass[lettersize,journal]{IEEEtran}
\usepackage{amsmath,amsfonts}
\usepackage{algorithmic}
\usepackage{algorithm}
\usepackage{array}
\usepackage[caption=false,font=normalsize,labelfont=sf,textfont=sf]{subfig}
\usepackage{textcomp}
\usepackage{stfloats}
\usepackage{url}
\usepackage{verbatim}
\usepackage{graphicx}
\usepackage{multirow}
\usepackage{threeparttable} 
\usepackage{booktabs}
\usepackage{cite}
\begin{document}

\title{IntentReact: Guiding Reactive Object-Centric Navigation \\via Topological Intent}


\author{Yanmei Jiao$^{1}$,
Anpeng Lu$^{1}$,
Wenhan Hu$^{1}$,  
Rong Xiong$^{2}$,
Yue Wang$^{2}$, 
Huajin Tang$^{3}$, and
Wen-an Zhang$^{4}$
\thanks{$^{1}$Yanmei Jiao, Anpeng Lu and Wenhan Hu are with the School of Information Science and Engineering, Hangzhou Normal University, Hangzhou, P.R. China. }%
\thanks{$^{2}$Rong Xiong and Yue Wang are with the State Key Laboratory of Industrial Control and Technology, Zhejiang University, Hangzhou, P.R. China.}
\thanks{$^{3}$Huajin Tang is with Neuromorphic Computing and Robotic Cognition Lab, Zhejiang University, Hangzhou, P.R. China.}
\thanks{$^{4}$Wen-an Zhang is with the Department of Automation, Zhejiang University of Technology, Hangzhou 310023, China.}
\thanks{Corresponding author: Wen-an Zhang ({\tt\small wazhang@zjut.edu.cn}). }
}


\markboth{Journal of \LaTeX\ Class Files,~Vol.~14, No.~8, August~2021}%
{Shell \MakeLowercase{\textit{et al.}}: A Sample Article Using IEEEtran.cls for IEEE Journals}


\maketitle

\begin{abstract}
Object-goal visual navigation requires robots to reason over semantic structure and act effectively under partial observability.
Recent approaches based on object-level topological maps enable long-horizon navigation without dense geometric reconstruction, but their execution remains limited by the gap between global topological guidance and local perception-driven control.
In particular, local decisions are made solely from the current egocentric observation, without access to information beyond the robot’s field of view, causing the robot to persist along its current heading even when initially oriented away from the goal and moving toward directions that do not decrease the global topological distance.
In this work, we propose \emph{IntentReact}, an intent-conditioned object-centric navigation framework that introduces a compact interface between global topological planning and reactive object-centric control.
Our approach encodes global topological guidance as a low-dimensional directional signal, termed \emph{intent}, which conditions a learned waypoint prediction policy to bias navigation toward topologically consistent progression.
This design enables the robot to promptly reorient when local observations are misleading, guiding motion toward directions that decrease global topological distance while preserving the reactivity and robustness of object-centric control.
We evaluate the proposed framework through extensive experiments, demonstrating improved navigation success and execution quality compared to prior object-centric navigation methods.
\end{abstract}


\section{Introduction}
\IEEEPARstart{V}{isual} navigation is a core capability for mobile robots operating in indoor environments.
Recent research has increasingly focused on \emph{object-goal navigation}, where a robot is tasked with reaching a target object category using onboard perception and semantic understanding.
In this setting, object-level topological representations have emerged as a promising abstraction, enabling long-horizon navigation by reasoning over semantic objects and their connectivity rather than dense geometric maps~\cite{savinov2018semi, chaplot2020neural,shah2023vint}.

\begin{figure}[t]
\centering
\includegraphics[width=\linewidth]{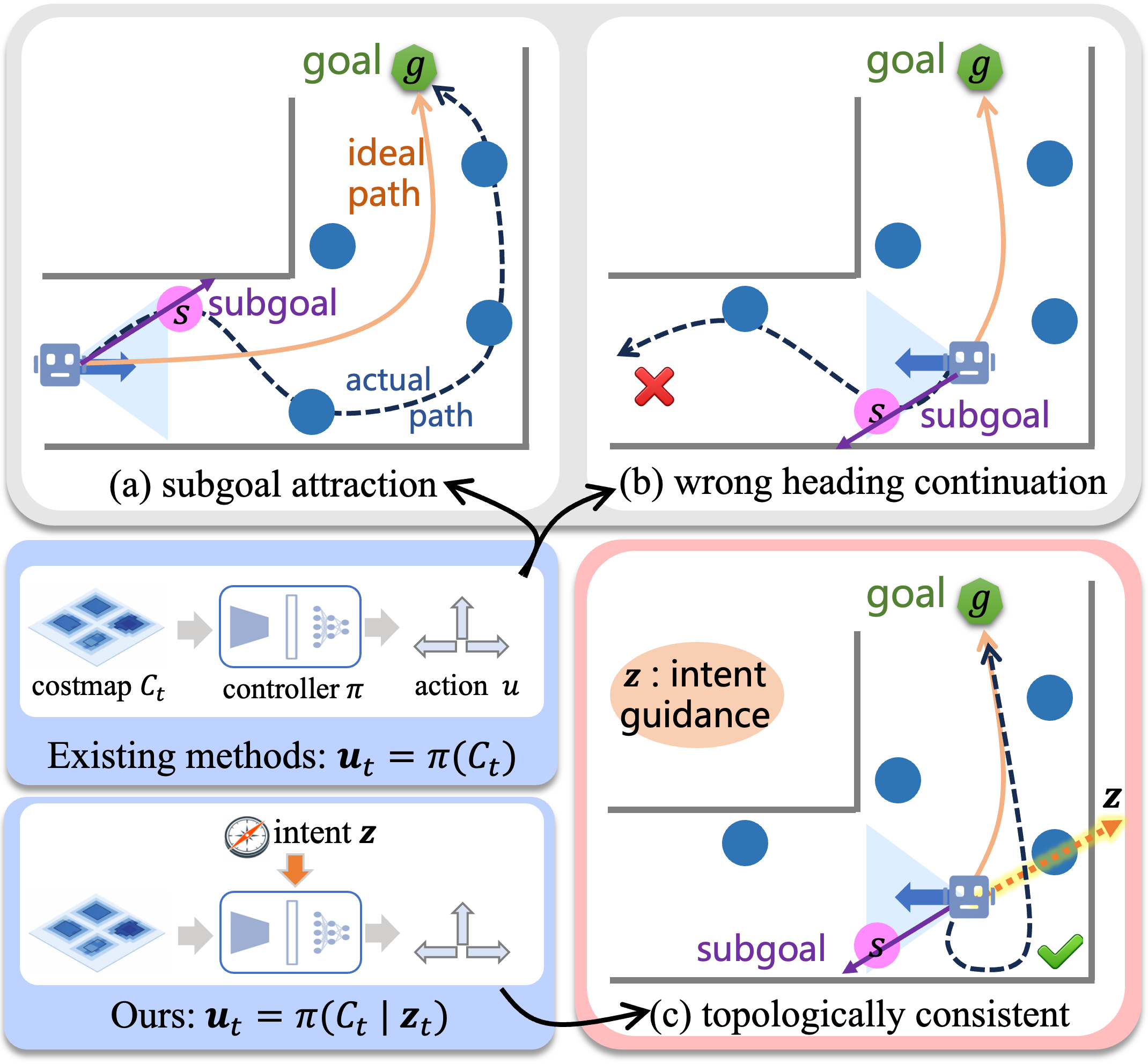}
\caption{Existing methods exhibit \emph{short-sighted} behavior due to the decoupling of global planning and local control, leading to locally reasonable yet globally inefficient or even incorrect trajectories: (a) excessive subgoal attraction results in zigzag trajectories and low overall efficiency; (b) large initial heading errors cause persistent wrong-direction motion. Our method introduces a low-dimensional \emph{intent} to condition the waypoint policy, enforcing topologically consistent progression and enabling efficient correction, as shown in (c).
}
\label{fig:teaser}
\end{figure}

Building on object-level topological maps, several navigation pipelines decompose the problem into global object-level planning and local reactive control.
A global planner estimates topological distances from objects to the goal, while a learned controller reacts to the set of object instances visible in the current egocentric observation~\cite{shah2023gnm, cai2024pixelguided,garg2025objectreact}.
This setting is particularly challenging for monocular robots with limited fields of view, where only a partial observation of the environment is available at each step.
This object-centric formulation reduces sensitivity to viewpoint and embodiment variations, supports open-set semantic queries, and has shown strong potential for visual navigation in complex environments.

Despite their effectiveness, existing object-centric navigation methods remain fundamentally constrained by partial observability.
At each time step, decisions are made based solely on the current observation and the costs of visible objects.
In environments with symmetric layouts, corridor-like structures, or when the robot is initially oriented away from the target, locally optimal object cues do not necessarily correspond to a decrease in the true topological distance.
As a result, the robot may continue moving along its current heading, inadvertently increasing its distance to the goal while appearing locally rational.
In essence, this ambiguity arises from a disconnect between global topological guidance and local decision-making: although object-level planning encodes long-horizon structure, its guidance is only indirectly reflected through the costs of currently visible objects, leaving the local controller without an explicit mechanism to favor globally consistent progression beyond the robot’s field of view.

Inspired by human navigation behavior, where a consistent sense of global direction is maintained to guide motion and enable timely correction even after local deviations, we argue that an explicit directional signal is essential for resolving such ambiguity. Building on this intuition, we introduce \emph{IntentReact}, an {intent-conditioned object-centric navigation} framework to bridge the gap between global topological guidance and local perception-driven control.
Our key idea is to compress global structural information into a low-dimensional directional signal, termed \emph{intent}, which provides guidance toward future topological progression without prescribing explicit actions, trajectories, or subgoals.
Concretely, intent represents the direction toward the 2-hop along the globally optimal object-level path, expressed in the robot’s local frame.
Intent introduces a directional bias over unobservable future structure, while leaving local action selection grounded in the robot’s current perceptual observation.

We incorporate intent into an object-level navigation policy by conditioning a learned waypoint predictor through feature-wise linear modulation (FiLM).
This design allows intent to modulate internal feature representations while preserving the controller’s strong grounding in local perceptual cues.
As a result, the policy gains the ability to promptly reorient toward the goal when facing misleading local optima—such as when initially oriented away from the target—while remaining responsive to immediate environmental constraints.
To further improve execution robustness, we introduce a feasibility-aware waypoint refinement module that refines predicted waypoints using bird’s-eye-view
(BEV) traversability constraints, forming a unified loop that combines learned decision-making with explicit feasibility reasoning.

Although our experiments focus on navigation with a known object-level topological map, the proposed framework is intentionally modular.
Intent can originate from a variety of sources, including human-provided directional instructions (e.g., ``go straight, then turn right''), semantic knowledge bases, or exploration heuristics.
In this sense, intent serves as a unifying interface between high-level guidance and low-level reactive control under partial observability.

In summary, this paper makes the following contributions:
\begin{itemize}
    \item We propose an \emph{intent-conditioned object-centric navigation} framework that explicitly introduces a global topological intent into a reactive object-level controller. 
    The intent is used as a conditioning signal to modulate the controller’s feature representation, effectively bridging global planning and local decision-making under partial observability.

    \item We introduce a compact \emph{2-hop intent representation} that captures the minimal global information required to resolve local ambiguity. 
    By encoding the direction of topological distance descent using only the next-hop along the planned path, the proposed representation achieves a balance between global consistency and local executability.

    \item We present a \emph{feasibility-aware waypoint refinement} module that enforces geometric consistency via BEV-based traversability reasoning. 
    This module projects predicted waypoints onto feasible regions, forming a unified closed loop between learning-based control and geometric feasibility.
\end{itemize}



\section{Related Work}
\subsection{Object-Goal Visual Navigation}

Object-goal visual navigation studies the problem of reaching a target object category using onboard perception and semantic understanding, rather than precise metric coordinates.
This formulation has been widely adopted in embodied navigation benchmarks such as ObjectNav and InstanceImageNav, which emphasize semantic reasoning and generalization across environments and embodiments~\cite{ramakrishnan2021hm3d, wani2020multion,krantz2022instance}.

Early work combines semantic perception with exploration and mapping to reason about unseen target locations~\cite{chaplot2020object, du2020objectrelation,yoo2024commonsense,kim2022topological}.
More recent approaches exploit object-level abstractions to reduce dependence on accurate localization and dense geometry, demonstrating improved robustness under visual variation and partial observability~\cite{savinov2018semi,shah2022viking}.
These results highlight the promise of object-centric representations as a stable interface for long-horizon navigation.

\subsection{Object-Centric Navigation with Topological Maps}
Early topological navigation represents nodes as image-based places, whereas recent approaches increasingly adopt object-level representations, where nodes correspond to segmented semantic objects to enable more structured navigation.
RoboHop constructs a segment-level topological map and enables efficient object-centric retrieval by reasoning over object connectivity, while employing a path-length-weighted pixel offset controller for local execution~\cite{garg2024robohop}.
ObjectReact further enriches this paradigm by building a relative 3D scene graph to capture object relationships, and learns an object-centric reactive controller that predicts future waypoints from object-centric costmaps, demonstrating strong generalization and sim-to-real transfer~\cite{garg2025objectreact}.

In parallel, several works highlight that topological guidance alone is insufficient for reliable execution.
TANGO addresses this limitation by incorporating depth and traversability constraints on top of topological goal maps, converting sub-goals into locally feasible trajectories via subgoal-driven control~\cite{podgorski2025tango}.
Related hybrid approaches similarly combine semantic planning with geometric or metric execution to ensure safety and feasibility~\cite{chaplot2020neural}.

Together, these methods expose a recurring disconnect between global topological guidance and local decision-making, motivating approaches that explicitly bridge planning and execution under partial observability.

\subsection{Learning-Based Navigation under Partial Observability}

Partial observability is a fundamental challenge in visual navigation, as agents must act based on incomplete and egocentric sensory input.
A common strategy is to equip policies with memory or belief representations that aggregate information over time, including recurrent neural networks, attention-based memory, or learned latent maps~\cite{gupta2017cognitive, parisotto2018neural,chaplot2020neural}.
These approaches aim to infer unobserved structure from observation history, but often rely on high-dimensional internal representations and tightly coupled policy architectures.

In contrast, object-centric navigation frameworks emphasize reactive control grounded in currently visible object cues, prioritizing simplicity, generalization, and sim-to-real transfer~\cite{garg2025objectreact}.
Our approach follows this line of work while introducing an explicit mechanism to incorporate global topological guidance into local decision-making.

\section{Method}

\begin{figure*}[t]
\centering
\includegraphics[width=0.9\linewidth]{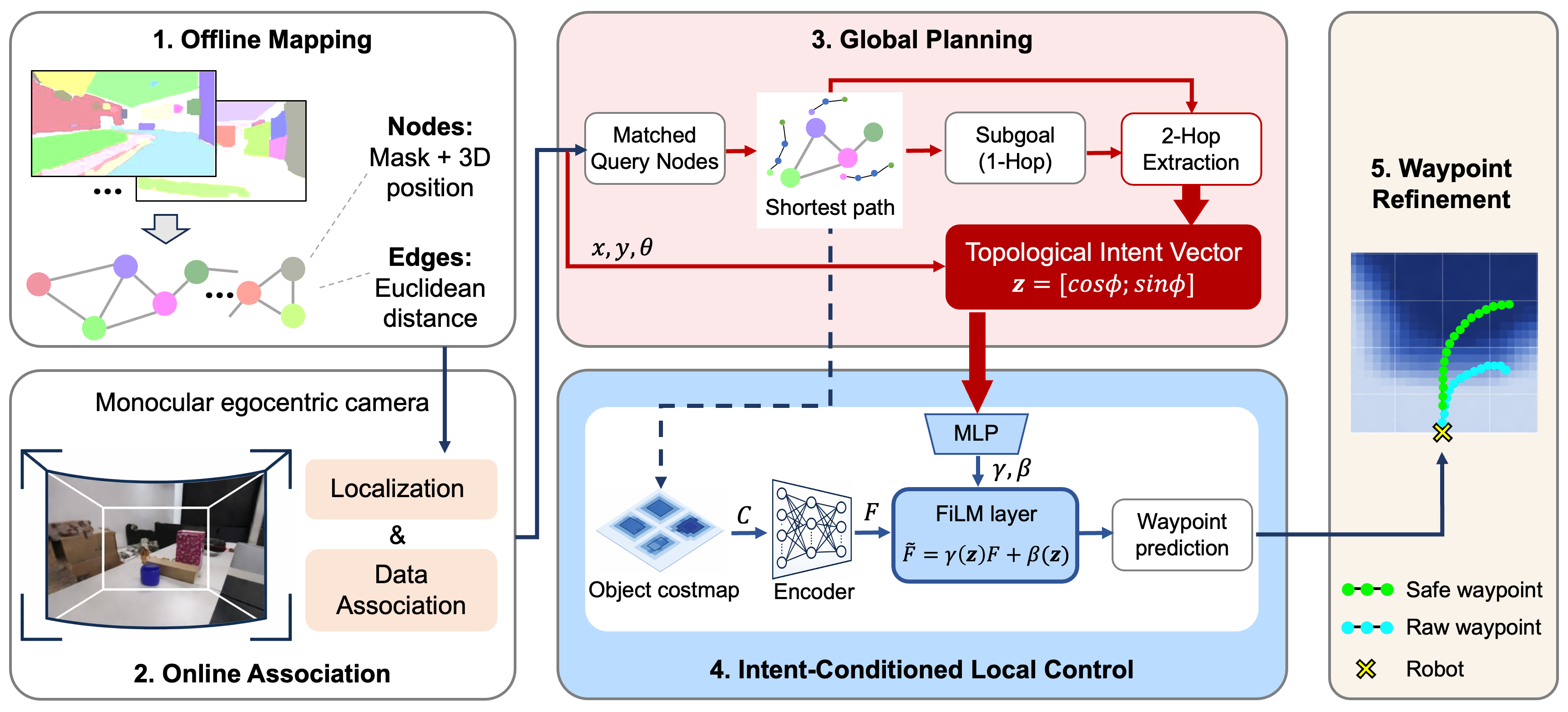}
\caption{IntentReact Navigation pipeline. (1) We construct an object-level topological map with 3D node information.
(2) During online execution, the current observation is matched to the map to obtain a set of query nodes.
(3) For each query node, global planning is performed to obtain a shortest path, from which an object-level costmap is built and an intent vector indicating the direction of decreasing topological distance is computed and fed into the local control stage.
(4) The intent is incorporated as a conditional signal to modulate the controller’s feature representation, enabling global guidance rather than hard intervention.
(5) A BEV costmap is further used to correct the feasibility of the predicted waypoint, forming a unified closed loop between learned control and geometric feasibility.}
\label{fig:overview}
\end{figure*}

Our method follows an object-centric navigation pipeline consisting of object-centric mapping, global object-level planning, and local reactive control \cite{garg2024robohop,chaplot2020object}. 
Unlike existing approaches that make control decisions solely based on the topological distances from objects visible in the current observation to the goal, our method additionally derives a compact directional signal from the global planning stage, termed \emph{intent}, which is used to condition the local controller. 
This signal biases local decision-making toward globally consistent progression while preserving reactivity to immediate perceptual cues. 
In the following subsections, we describe the mapping, planning, and control components of the pipeline, as also shown in Fig. \ref{fig:overview}.

\subsection{Object-Centric Topological Mapping}
We represent the object topological map as a graph $G = (N, E)$, where $N$ denotes the set of object nodes in the map and $E$ denotes the connectivity edges between objects.

\textbf{Object Nodes.}
For each RGB observation obtained during mapping, we extract open-set object instances using foundation models such as SAM2\cite{ravi2025sam2} or FastSAM\cite{zhao2023fast}. Each object instance is represented as a node $n_i \in N$.
The attributes of a node include a 2D binary segmentation mask $M_i$ and a 3D global coordinate $\mathbf{p}_i$.
And $\mathbf{p}_i$ is obtained by projecting the centroid of the instance mask into the global frame using the estimated camera pose and depth.
The camera pose can be obtained either from standard monocular visual odometry pipelines~\cite{mur2017orbslam2,engel2018dso} or from offline feed-forward 3D reconstruction models that jointly recover camera poses and scene geometry from image collections~\cite{wang2025vggt,keetha2025mapanything}, while depth is obtained from monocular depth estimation\cite{yang2024depthanything}.

\textbf{Edges.} Edges encode connectivity between objects, capturing both spatial relationships within a single observation and temporal relationships across observations. 
Specifically, intra-image edges are constructed between object instances co-visible in the same image.
These edges encode the Euclidean distance $e_{ij}$ between object pairs, which are identified using a Delaunay triangulation over the object centers.

Inter-image edges are established by tracking objects across different time steps, reflecting temporal continuity along the agent trajectory.
Concretely, we perform pixel-level matching using SuperPoint~\cite{detone2018superpoint} and LightGlue~\cite{lindenberger2023lightglue}, and then associate object instances based on the overlap between matched pixels and instance masks.
For such edges, we set $e_{ij} = 0$ to indicate that the connected nodes correspond to the same physical object.
This prevents additional path costs from being introduced between temporally repeated instances of the same object during planning.



\subsection{Global Object-level Planning}\label{sec:intent-def}
Given an observation image with a target object prompt, navigation proceeds iteratively.
At each step, we first retrieve a set of associated map images using place recognition.
Using the instance matching method described earlier, each query object node in the current observation is associated with a corresponding map object node.

Based on the topological graph $G=(N,E)$, the global object-level planner computes the shortest path from each query node $n_q$ to the goal node $n_g$ using Dijkstra's algorithm.
Unlike ObjectReact, which uses these path-length values to construct an object-centric costmap as the controller input, our method additionally derives a low-dimensional intent vector during global planning.
The intent encodes the direction of decreasing global topological distance and is used together with the costmap to guide the controller, as described in the following section.

\textbf{Design Rationale.}
While the object-centric costmap encodes the relative desirability of visible objects, it does not indicate how the global topological distance is expected to evolve beyond the current field of view. 
To address this limitation, we introduce \emph{intent} as a compact directional signal that summarizes future topological progression.

Several alternatives could convey global guidance, such as encoding the full map, the entire planned path, or the direction toward the final goal. 
However, representing the full map or path introduces high-dimensional structure that is difficult to integrate into a lightweight reactive controller and unnecessarily couples the policy to a specific planning representation.
A simpler alternative is to use the direction toward the final goal. 
Yet this can be misleading when the globally optimal route initially deviates from the goal direction, such as when navigating around obstacles or through corridors, potentially causing premature turns or locally inconsistent actions.

\begin{figure}[t]
\centering
\includegraphics[width=\linewidth]{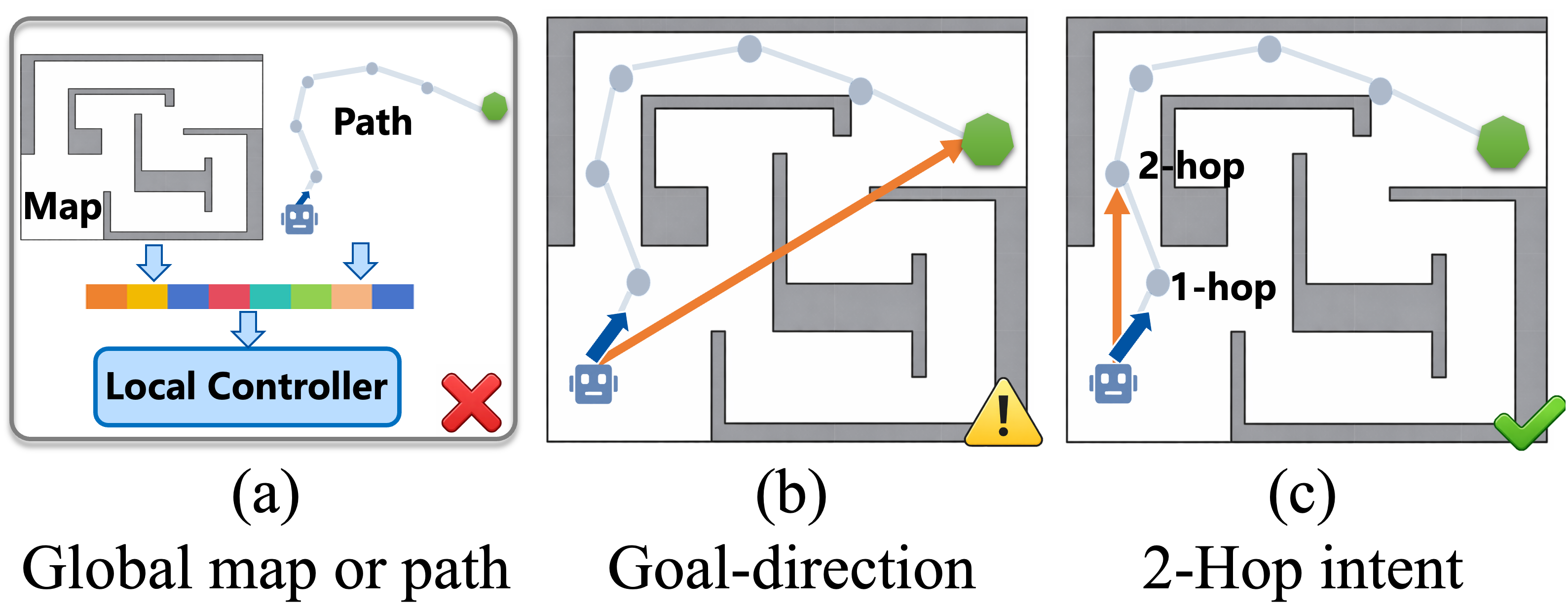}
\caption{Comparison of different conditioning strategies. 
(a) Global path conditioning leverages full map or planned path information, but introduces redundancy and weakens local reactivity. 
(b) Goal-direction conditioning provides a simple directional cue, but may lead to premature turning and local collisions. 
(c) The proposed 2-hop intent conditioning encodes a compact topological signal, effectively balancing global consistency and local feasibility.}
\label{fig:intent}
\end{figure}

Instead, we derive intent from the 2-hop node along the global object-level path, as illustrated in Fig. \ref{fig:intent}. 
The 2-hop captures the immediate direction of decreasing topological distance while remaining consistent with the global plan, providing sufficient guidance to resolve local ambiguity while preserving reactive control.

\textbf{Intent Definition.}
At each step, the current observation is associated with a set of matched map nodes 
${N}_q \subset N$ through instance matching.
We select the \emph{sub-goal node}
\begin{equation}
n_{\text{sub}} = \arg\min_{n \in {N}_q} d(n),
\end{equation}
where $d(n)$ denotes the topological distance from node $n$ to the goal node $n_g$.
Let
\begin{equation}
\mathcal{P}(n_{\text{sub}}, n_g)
=
\{n_{\text{sub}}, n_1, \dots, n_g\}
\end{equation}
denote the shortest path to the goal.
Because multiple nodes may correspond to the same physical object across different observations, consecutive nodes along the path may share identical topological distance values.
We therefore define the \emph{2-hop node} (denoted as $n_{\text{next}}$) as the first node along the path whose distance strictly decreases:
\begin{equation}
n_{\text{next}}
=
\min_{n \in \mathcal{P}(n_{\text{sub}}, n_g)}
\{n \mid d(n) < d(n_{\text{sub}})\}.
\end{equation}

Let $(x_{\text{next}},y_{\text{next}})$ denote the planar coordinates obtained by projecting $\mathbf{p}_{\text{next}}$ onto the ground plane, and let $(x_t,y_t)$ and $\theta_t$ denote the robot's planar position and yaw.
Intent is defined as
\begin{equation}
\mathbf{z} =
\begin{bmatrix}
\cos\phi \\
\sin\phi
\end{bmatrix},
\quad
\phi =
\mathrm{atan2}(y_{\text{next}}-y_t,\;x_{\text{next}}-x_t)-\theta_t.
\end{equation}

\subsection{Intent-Conditioned Local Control}

We incorporate the intent signal into the object-centric controller to guide local decisions toward globally consistent topological progress while preserving reactive execution. 
The controller follows the waypoint prediction architecture of ObjectReact\cite{chaplot2020object} and augments it with intent-based feature conditioning.

\textbf{Object-Centric Costmap.}
For each visible object node $n_q$, the global planner provides its topological distance $d(n_q)$ to the goal node $n_g$. 
These scalar distances are embedded using a fixed sinusoidal encoding and projected onto the image plane through the corresponding instance masks, producing a dense object-centric costmap. 
The costmap is processed by a goal encoder to obtain a feature map
\[
F \in \mathbb{R}^{C\times H\times W}.
\]
This representation follows the object costmap design in ObjectReact, which avoids explicit RGB inputs and has been shown to improve generalization to unseen environments and sim-to-real transfer.

\textbf{Intent Conditioning.}
To introduce global guidance, we condition the controller on the intent vector $\mathbf{z}$ using feature-wise linear modulation (FiLM) \cite{perez2018film}. 
A small multilayer perceptron maps $\mathbf{z}\in\mathbb{R}^2$ to channel-wise modulation parameters $(\boldsymbol{\gamma},\boldsymbol{\beta})\in\mathbb{R}^C$, and the feature map is modulated as
\begin{equation}
\tilde{F}_{c,h,w}=\boldsymbol{\gamma}_c(\mathbf{z})F_{c,h,w}+\boldsymbol{\beta}_c(\mathbf{z}).
\end{equation}

FiLM is inserted after the goal encoder and before global pooling, allowing the intent signal to influence intermediate representations while preserving spatial structure. 
To maintain compatibility with the pretrained controller, the FiLM layer is initialized close to an identity transformation ($\boldsymbol{\gamma}\approx1$, $\boldsymbol{\beta}\approx0$), so the policy initially behaves similarly to the original model and gradually learns to exploit the intent signal during training.

\textbf{Training Data and Setup.}
We train the controller using the Habitat-Matterport3D (HM3Dv0.2) dataset\cite{ramakrishnan2021hm3d} with the InstanceImageNav (IIN-HM3D-v3) benchmark\cite{krantz2022instance}. 
Shortest paths between start and goal states are obtained from the simulator using geodesic distance and converted into waypoint trajectories through interpolation of translational and rotational motions.

During training, object instances and depth from the simulator are used to construct object-centric costmaps and topological distances for each observation. 
The intent vector is computed from the global object-level plan by identifying the 2-hop node along the shortest path toward the goal and converting its direction into a unit vector in the robot frame.

The controller is optimized with the same waypoint prediction objective used for training the base policy. 
To stabilize learning, we adopt a staged training strategy where the FiLM parameters are first trained while the rest of the controller remains frozen, followed by joint fine-tuning of the full network.

\subsection{Feasibility-Aware Waypoint Refinement}

Although the intent-conditioned controller predicts locally consistent waypoints, learning-based policies may occasionally generate trajectories that intersect obstacles or lie close to non-traversable regions. 
To improve execution robustness, we introduce a lightweight waypoint refinement module that enforces local geometric feasibility at execution time.

\textbf{BEV Traversability Map.}
We assume access to a local BEV traversability map centered at the robot that encodes reachable free space. 
Such a representation can be obtained from depth sensing or monocular depth estimation with geometric projection. 
The BEV map is used only to evaluate the feasibility of predicted waypoints and does not participate in planning or policy learning.

\textbf{Waypoint Refinement.}
Given a predicted waypoint $\mathbf{w}$ in the robot frame, we project it onto the BEV map to verify traversability. 
If $\mathbf{w}$ lies in free space, it is executed directly. 
Otherwise, the waypoint is projected to the nearest feasible location along its direction or adjusted within a local neighborhood that satisfies traversability constraints. 
This operation preserves the intended motion direction while ensuring physical feasibility.

\textbf{Execution Role.}
The refinement module operates deterministically and does not modify the learned controller or its internal representations. 
It therefore acts as a lightweight safety layer that complements the intent-conditioned policy. 
This separation allows the policy to focus on semantic and topological reasoning, while geometric feasibility is enforced during execution.


\begin{table*}[t]
\centering
\caption{Performance comparison under the \textbf{GT-Topological} setting. }
\label{tab:gttopo}
\setlength{\tabcolsep}{4pt}
\begin{tabular}{lcccccccccccc}
\toprule
{\multirow{3}{*}{Method}} 
& \multicolumn{3}{c}{{Imitate}} 
& \multicolumn{3}{c}{{Alt Goal}} 
& \multicolumn{3}{c}{{Shortcut}} 
& \multicolumn{3}{c}{{Reverse}} \\
\cmidrule(lr){2-4} \cmidrule(lr){5-7} \cmidrule(lr){8-10} \cmidrule(lr){11-13}
 & SR & SPL & SSPL 
 & SR & SPL & SSPL 
 & SR & SPL & SSPL 
 & SR & SPL & SSPL \\
\midrule
RoboHop\cite{garg2024robohop} 
& 54.63 & 53.13 & 67.72 
& 21.30 & 20.63 & 35.32 
& 27.62 & 25.98 & 45.12 
& 38.89 & 34.89 & 49.82 \\

TANGO\cite{podgorski2025tango} 
& 64.81 & 59.21 & 72.67 
& 24.07 & 20.57 & 36.85 
& 34.29 & 30.56 & 48.90 
& 55.56 & 48.48 & 62.28 \\

ObjectReact-ft\cite{garg2025objectreact} 
& 63.89 & 57.84 & 76.05 
& 26.85 & 23.79 & 38.50 
& 39.05 & 34.55 & 57.28 
& 51.85 & 48.04 & 64.94 \\

{(Ours) IntentReact} 
& \textbf{81.48} & \textbf{76.12} & \textbf{87.85} 
& \textbf{34.26} & \textbf{31.07} & \textbf{43.69} 
& \textbf{60.95} & \textbf{53.86} & \textbf{74.35} 
& \textbf{62.04} & \textbf{59.89} & \textbf{70.18} \\
\bottomrule
\end{tabular}
\end{table*}

\section{Experimental Results}
\textbf{Dataset.} 
We evaluate our method on the HM3Dv0.2 benchmark \cite{ramakrishnan2021hm3d} using the validation split of the IIN-HM3D-v3 task\cite{krantz2022instance}. The evaluation set contains 36 indoor environments. For each scene, we sample three navigation episodes with distinct object goals, resulting in a total of 108 episodes. For each episode, we generate a reference trajectory using the simulator’s geodesic planner. This trajectory is used to construct the prior map during an offline mapping phase. During evaluation, to ensure meaningful navigation difficulty, the start and goal locations in all episodes are separated by at least 5\,m in geodesic distance. Map images are generated assuming a fixed sensor height of 1.3\,m. 
During evaluation, the agent operates with a robot height of 0.4\,m, consistent with the ObjectReact baseline. For fair comparison, all methods are evaluated using the same map across all experiments.

\textbf{Metrics.}
An episode is considered successful if the agent reaches within 1\,m of the goal within a maximum of 300 steps, using an oracle stopping condition\cite{yadav2023habitat}. We report Success Rate (SR), Success weighted by Path Length (SPL)\cite{anderson2018evaluation}, and Soft-SPL (SSPL)\cite{datta2021egocentric}, which measure task completion, trajectory efficiency, and progress toward the goal. Additionally, we report the average number of steps among successful episodes to further assess navigation efficiency.

\subsection{Benchmark Evaluation with GT Perception}
We first evaluate navigation performance under the \textit{GT-Topological} setting, where segmentation, depth, and localization are all provided by simulator ground truth, as in \cite{podgorski2025tango}. 
This setup isolates the role of planning and control while assuming perception is solved.
We evaluate navigation under four task settings used in ObjectReact baseline: \textit{Imitate}, \textit{Alt Goal}, \textit{Shortcut}, and \textit{Reverse}. 
These tasks respectively test trajectory following, reaching a previously observed but unvisited goal, taking a shorter route to the goal, and traversing the trajectory in the opposite direction. All tasks are evaluated over 108 episodes and we report the mean performance across episodes. The only exception is \textit{Shortcut}, which contains 105 episodes due to three missing task instances.

Tab.~\ref{tab:gttopo} reports results across four tasks. Since the original ObjectReact release provides only the prebuilt maps but not the full map construction pipeline, we reconstruct the maps ourselves for all methods to ensure a fair comparison. 
For learning-based methods, we finetune the ObjectReact policy on the reconstructed maps until it reaches performance comparable to the original model on its native maps. This variant is denoted as \textit{ObjectReact-ft}. Results show that our method consistently outperforms all baselines across the four tasks.
The improvements are most significant in the \textit{Imitate} and \textit{Shortcut} settings, where our approach improves SPL by 17.0 and 19.3 points respectively compared to the strongest baseline.
These gains indicate that the proposed intent guidance helps the controller maintain globally consistent navigation decisions, particularly when multiple locally feasible actions exist.

\begin{figure*}[t]
\centering
\includegraphics[width=\linewidth]{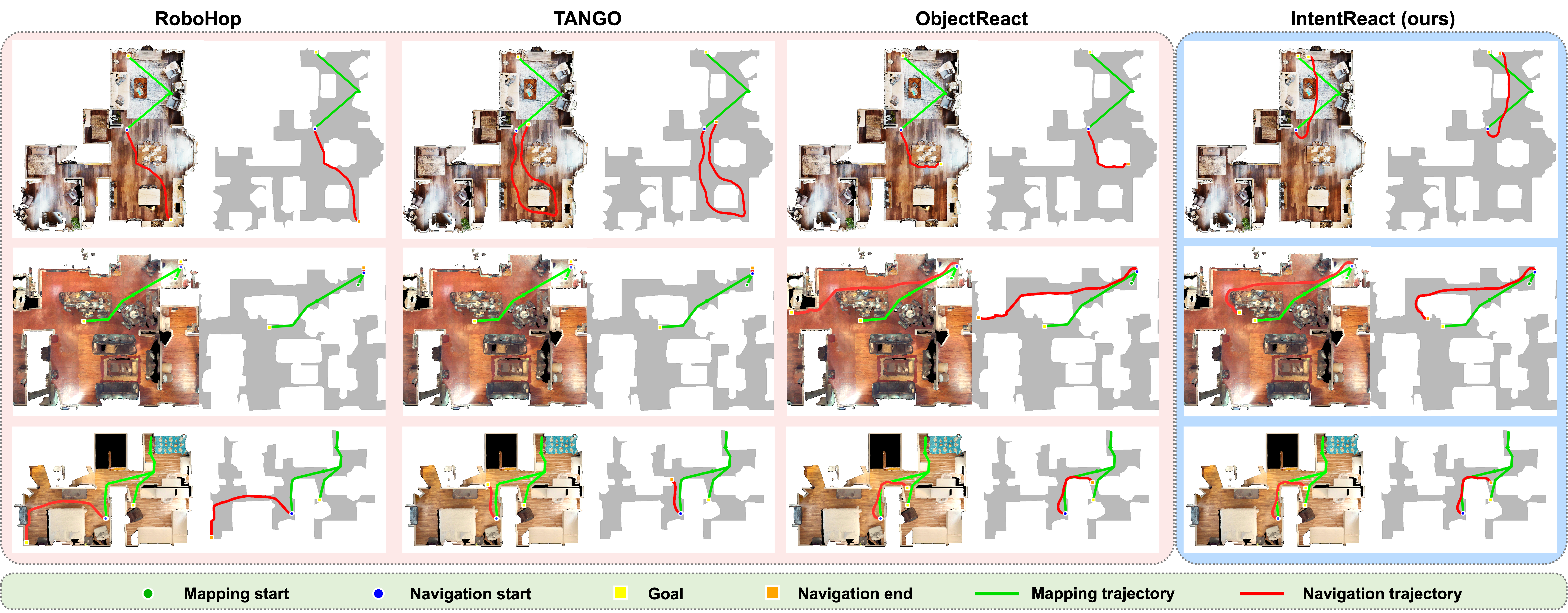}
\caption{Qualitative results under the \textit{Shortcut} setting. When the initial heading deviates from the goal direction, RoboHop, TANGO, and ObjectReact often follow misleading local cues, resulting in inefficient or failed trajectories. In contrast, our method leverages the intent signal to correct early deviations and maintain topologically consistent progression. In cases without ambiguity, it behaves identically to ObjectReact, indicating that intent acts as a soft bias without disrupting the underlying reactive policy.}
\label{fig:case}
\end{figure*}

\begin{figure}[t]
\centering
\includegraphics[width=\linewidth]{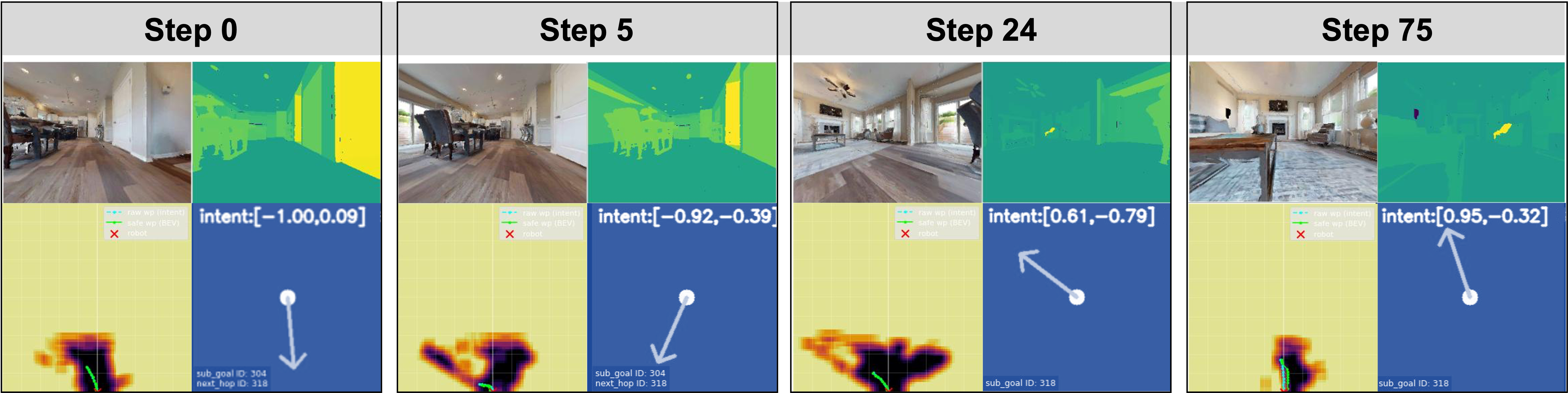}
\caption{Step-by-step visualization of our method for the first case in Fig.~\ref{fig:case}. 
At each step, four panels show the RGB observation, segmentation, BEV waypoint, and intent direction. 
Despite a large initial heading error, intent provides consistent directional bias, enabling rapid correction toward the goal (step 24). 
BEV refinement further ensures safe execution by projecting raw waypoints onto feasible regions (step 75).}
\label{fig:step}
\end{figure}

Fig.~\ref{fig:case} presents representative qualitative results under the \textit{Shortcut} setting, where the mapped trajectory is intentionally suboptimal and the agent is expected to discover a more efficient path, providing a strong test of navigation performance. 
We visualize three typical cases. 
In the first two cases, the initial heading of the agent significantly deviates from the optimal direction toward the goal. 
As a result, RoboHop and TANGO tend to follow the initial heading and move further away from the goal. 
Although TANGO eventually turns back in the first case after reaching a dead end, it exceeds the step limit and fails to reach the target. 
ObjectReact also deviates from the goal in the first case. 
In the second case, it initially turns toward the correct direction, but lacks consistent guidance near the goal and ultimately drifts away, resulting in a low SSPL.

In contrast, our method benefits from the global topological descent direction encoded by the intent signal. 
In the first case, it corrects the wrong initial heading early and produces a more efficient trajectory than the mapping path. 
In the second case, unlike ObjectReact, it successfully reaches the goal without drifting.
In the third case, our trajectory is identical to that of ObjectReact. 
This indicates that the intent signal only acts when directional ambiguity arises and does not interfere with the original reactive control policy. 
In other words, intent serves as a soft bias rather than a hard constraint, preserving the strengths of local reactive control while improving global consistency.

Fig.~\ref{fig:step} presents a step-by-step visualization of our method for the first case in Fig.~\ref{fig:case}. 
At each step, the four panels correspond to the RGB observation, semantic segmentation, BEV-based waypoint, and intent direction, respectively.
As shown in Fig.~\ref{fig:step}, due to the large initial heading deviation, the local controller initially predicts forward-facing waypoints (step 0). 
However, the intent signal consistently provides a backward directional bias, gradually correcting the waypoint orientation. 
Within only 5 steps, the predicted waypoints start to turn left, and the agent continues to reorient accordingly. 
By step 24, the agent has already turned toward a direction where the target object becomes visible.
Step 75 illustrates a typical case of BEV-based refinement. 
When encountering obstacles, the raw waypoint is successfully projected onto a feasible region, resulting in a safe waypoint for execution. 

\subsection{Robustness to Initial Orientation Ambiguity}

To further evaluate the role of the proposed intent guidance, we introduce an additional task, \textit{Opposite}. 
Unlike \textit{Reverse}, which changes the traversal direction, \textit{Opposite} keeps the same route but perturbs the agent’s initial heading. 
Specifically, the initial orientation is rotated around the vertical axis by a predefined angle $\{0^\circ,60^\circ,120^\circ,150^\circ,180^\circ\}$ with a random sign per episode. 
This setting introduces increasing directional ambiguity at the start of navigation and tests whether the controller can recover globally consistent motion.

\begin{figure}[btp]
\centering
\includegraphics[width=\linewidth]{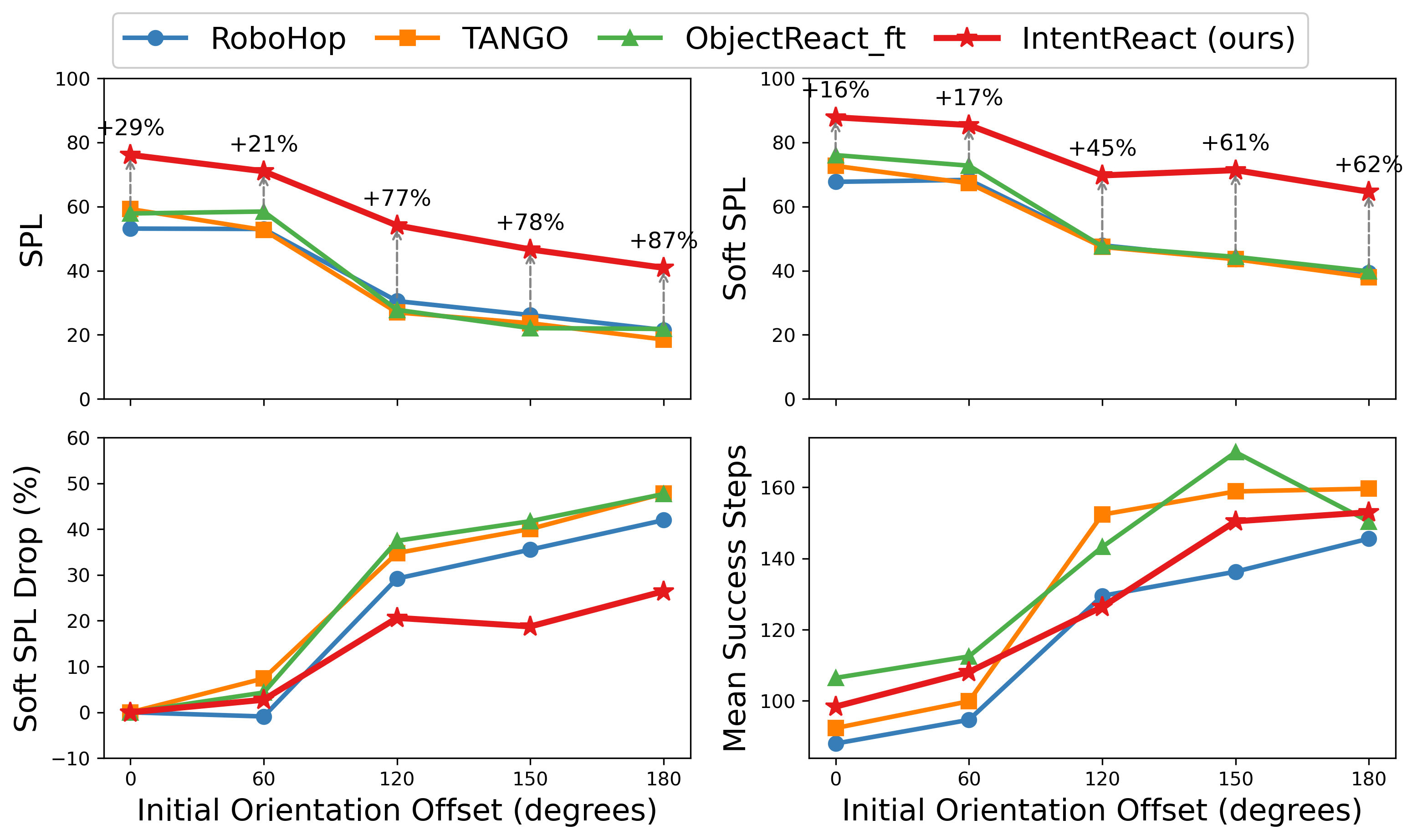}
\caption{Performance under different initial orientation offsets in the \textit{Opposite} task. Annotated percentages indicate the relative improvement over the strongest baseline.}
\label{fig:opposite}
\end{figure}

The experiment is conducted under the \textit{GT-Topological} setting to isolate the effect of control decisions. 
Fig.~\ref{fig:opposite} reports the resulting SPL, SSPL, SSPL drop percentage relative to the $0^\circ$ case, and the mean number of steps among successful episodes. 
The SSPL drop is computed as
\[
\text{Drop}(\theta)=\left(1-\frac{\text{SSPL}(\theta)}{\text{SSPL}(0)}\right)\times100.
\]
The annotated percentages indicate the relative improvement of our method over the strongest baseline at each offset, while the arrows highlight the corresponding performance gaps.

As the initial orientation offset increases, navigation becomes more challenging due to the growing mismatch between the agent's initial heading and the optimal navigation direction. 
As shown in Fig.~\ref{fig:opposite}, our method consistently achieves higher SPL and SSPL than all baselines across all offsets. 
Notably, the performance gain over the strongest baseline increases as the offset grows, indicating that the benefit of intent guidance becomes more pronounced under larger orientation perturbations. 
At the extreme $180^\circ$ offset, the SPL improvement reaches up to $87\%$ relative to the best baseline, demonstrating the strong corrective capability provided by the proposed intent signal. 
The SSPL drop curves further confirm this trend: while all baselines exhibit rapidly increasing degradation as the offset grows, our method maintains a substantially smaller drop and a smoother degradation trend. 

In addition, the SPL and SSPL curves show that our method maintains significantly higher navigation efficiency across all offsets. 
Although the mean number of steps increases with larger orientation errors for all methods, the increase for our approach remains moderate compared to the baselines. 
Overall, these results demonstrate that the intent-conditioned controller is substantially more robust to initial orientation ambiguity, enabling faster recovery toward globally consistent navigation trajectories.

\subsection{Sensitivity to Intent Estimation Errors}

We evaluate the robustness of the policy to imperfect intent predictions by introducing systematic angular noise to the intent direction. 
For each episode, we sample a constant bias $\epsilon \sim \mathcal{U}(-\alpha,\alpha)$ and modify the intent direction as $\theta'=\theta+\epsilon$. 
The bias remains fixed throughout the episode, simulating consistent directional errors that may arise from imperfect global planning or localization. 
We evaluate five noise levels $\alpha \in \{0^\circ,5^\circ,10^\circ,20^\circ,30^\circ\}$ under the \textit{GT-Topological} setting with \textit{Imitate} task.

Tab.~\ref{tab:intent_noise} shows that the policy remains stable under moderate angular noise. 
For biases up to $20^\circ$, the navigation performance remains almost unchanged, indicating that the policy does not rely on precise intent estimation. 
Even with a $30^\circ$ bias, SPL retains about $95\%$ of the original performance.

Interestingly, performance does not monotonically decrease with increasing noise. 
For example, small biases (e.g., $5^\circ$) even lead to slightly higher SPL than the noise-free case. 
This behavior suggests that the policy does not rigidly follow the intent direction but instead integrates it with visual observations during navigation. 
When the intent error becomes larger, the controller can rely more on local visual cues and object-centric signals to correct the trajectory.

Overall, these results indicate that the proposed intent representation serves as a \emph{soft guidance signal} rather than a hard geometric constraint, enabling the policy to remain robust under imperfect intent estimation.

\subsection{Ablation Study}

To analyze the contribution of each component in our framework, we perform a series of ablation experiments summarized in Tab.~\ref{tab:ablation}. 
We focus on three aspects: the effect of intent conditioning, different intent representations, and the BEV-based feasibility constraint.

\begin{table}[tbp]
\centering
\begin{threeparttable}
\caption{Sensitivity to intent estimation errors.}
\label{tab:intent_noise}
\setlength{\tabcolsep}{6pt}
\begin{tabular}{lcccc}
\toprule
Method & SR (\%) & SPL & SSPL & Retention \\
\midrule
(Base) ObjectReact-ft  & 72.22 & 66.54 & 83.45 & -- \\
\midrule
(Ours) IntentReact ($0^\circ$)  & 81.48 & 76.12 & 87.85 & 1.00 \\
(Ours) IntentReact ($5^\circ$)  & 82.41 & 77.23 & 88.11 & 1.01 \\
(Ours) IntentReact ($10^\circ$) & 81.48 & 75.70 & 88.08 & 0.99 \\
(Ours) IntentReact ($20^\circ$) & 79.63 & 75.24 & 87.14 & 0.99 \\
(Ours) IntentReact ($30^\circ$) & 77.78 & 72.42 & 85.98 & 0.95 \\
\bottomrule
\end{tabular}
\begin{tablenotes}
\footnotesize
\item Note: Retention denotes $\text{SPL} / \text{SPL}(0^\circ)$.
\end{tablenotes}

\end{threeparttable}
\end{table}

\begin{table}[tbp]
\centering
\caption{Ablation study of the proposed components.}
\label{tab:ablation}
\setlength{\tabcolsep}{6pt}
\begin{tabular}{lcccc}
\toprule
Method & Steps & SR (\%) & SPL & SSPL \\
\midrule
(Base) ObjectReact-ft & 106.46 & 63.89 & 57.84 & 76.05 \\
\midrule
(Ours) Base + Intent & 102.79 & 72.22 & 66.54 & 83.45 \\
(Ours) Base + Intent (sign) & 101.08 & 70.37 & 66.16 & 81.48 \\
(Ours) Base + Intent (concat) & 101.89 & 70.37 & 65.85 & 82.60 \\
(Ours) Base + Intent (+dist) & 103.20 & 70.37 & 65.42 & 81.55 \\
(Ours) Base + BEV & 104.38 & 75.00 & 68.45 & 80.43 \\
\midrule
(Ours) IntentReact & \textbf{98.40} & \textbf{81.48} & \textbf{76.12} & \textbf{87.85} \\
\bottomrule
\end{tabular}
\end{table}

\textbf{Effect of Intent Conditioning.}
We first introduce the intent signal into the baseline controller (Base). 
The intent vector is computed following the formulation described in Sec.~\ref{sec:intent-def}.
Conditioning the policy on this directional cue significantly improves navigation performance. 
As shown in Tab.~\ref{tab:ablation}, adding intent increases SPL from 57.84 to 66.54 and SSPL from 76.05 to 83.45, while also reducing the average trajectory length. 
These results suggest that providing an explicit directional bias helps the controller resolve navigation ambiguities and make more consistent decisions.

\textbf{Effect of Intent Representation.}
We further investigate whether the specific encoding of intent affects policy performance by evaluating several variants.

\emph{Intent sign} keeps only the sign of $\cos\theta$ and $\sin\theta$, preserving quadrant-level directional information while discarding angular magnitude. 
This variant tests whether coarse directional guidance alone is sufficient.

\emph{Intent concat} injects the intent vector by direct feature concatenation with the costmap representation instead of FiLM-style conditioning. 
This design allows us to examine whether the performance gain mainly comes from the intent signal itself or from the specific conditioning mechanism.

\emph{Intent dist} augments the intent representation with distance information. 
The distance term is computed as the shortest-path length on the topological graph, representing the remaining distance from the current sub-goal node to the final goal. 
This variant evaluates whether explicit distance magnitude provides additional useful information for the policy.

Interestingly, all variants achieve very similar performance, with SPL values ranging from 65.42 to 66.54. 
This observation suggests that the controller policy primarily relies on coarse directional guidance; precise angular information provides a modest improvement, while additional distance information has limited impact.
In other words, the policy mainly benefits from knowing the general direction toward the goal rather than exact geometric quantities.

\textbf{Effect of BEV Feasibility Constraint.}
We further evaluate a variant that enables the BEV-based feasibility module without intent guidance. 
This module projects predicted waypoints onto a traversable region estimated from the BEV costmap, enforcing geometric feasibility of the trajectory. 
The BEV constraint alone improves SPL to 68.45, demonstrating its effectiveness in reducing infeasible motions. 
However, the improvement in SSPL is smaller than that achieved by intent conditioning, since BEV primarily ensures local geometric feasibility but does not provide high-level directional guidance.

\textbf{Full Model.}
Combining intent conditioning with the BEV feasibility constraint yields the best performance across all metrics, achieving 76.12 SPL and 87.85 SSPL. 
In addition, the average trajectory length is significantly reduced, with the mean number of steps also decreasing, indicating more efficient navigation. 
These results highlight the complementary roles of the two modules: the intent signal provides global directional guidance along the topological path, while the BEV constraint enforces local geometric feasibility of the predicted motion.

\begin{table*}[t]
\centering
\caption{Navigation performance under the learned-perception setting. }
\label{tab:learned_perception}
\setlength{\tabcolsep}{4pt}
\renewcommand{\arraystretch}{1.15}
\begin{tabular}{clcccccccccccc}
\toprule
 & \multicolumn{1}{l}{\multirow{3}{*}{Method}} 
& \multicolumn{3}{c}{Imitate}
& \multicolumn{3}{c}{Alt Goal}
& \multicolumn{3}{c}{Shortcut}
& \multicolumn{3}{c}{Reverse} \\
\cmidrule(lr){3-5} \cmidrule(lr){6-8} \cmidrule(lr){9-11} \cmidrule(lr){12-14}
 &  & SR & SPL & SSPL & SR & SPL & SSPL & SR & SPL & SSPL & SR & SPL & SSPL \\
\midrule
\multirow{4}{*}{GT PR}
& RoboHop         & 48.15 & 43.70 & 55.90 & 19.44 & 14.90 & 27.23 & 17.14 & 13.74 & 27.10 & 12.96 & 10.48 & 19.84 \\
& TANGO           & 48.15 & 41.74 & 52.31 & 24.07 & 18.17 & 29.66 & 22.86 & 17.40 & 34.03 & 31.48 & 26.40 & 38.06 \\
& ObjectReact-ft  & 37.04 & 32.01 & 43.37 & 13.89 & 11.54 & 22.67 &  4.76 &  4.32 & 22.48 & 19.44 & 15.97 & 31.26 \\
& (Ours) IntentReact & \underline{64.81} & \textbf{60.49} & \textbf{68.46} & \textbf{28.70} & \textbf{23.56} & \textbf{37.36} & \textbf{39.05} & \textbf{32.83} & \textbf{51.13} & \underline{35.19} & \underline{28.16} & \underline{45.98} \\
\midrule
\multirow{1}{*}{NetVLAD}
& (Ours) IntentReact & \textbf{65.74} & \underline{58.86} & \underline{67.71} & \underline{25.93} & \underline{20.82} & \underline{37.34} & \underline{32.38} & \underline{27.54} & \underline{48.79} & \textbf{39.81} & \textbf{34.44} & \textbf{51.31} \\
\bottomrule
\end{tabular}
\end{table*}

\subsection{Evaluation with Learned Perception}

We further evaluate our method under a learned perception setting to better approximate real-world deployment. 
Specifically, we replace simulator-provided ground-truth perception with learned modules: FastSAM\cite{zhao2023fast} is used for object segmentation during both mapping and navigation, and SuperPoint + LightGlue are used for data association.
For localization during navigation, we consider two variants. 
First, we use ground-truth place recognition (GT-PR), where the agent retrieves the closest map image based on its 2D position, following ObjectReact. 
Second, we adopt a fully learned setting using NetVLAD\cite{arandjelovic2016netvlad} for place recognition (PR), which relies solely on visual features without access to ground-truth localization.

To compute the intent direction, an estimate of the robot’s yaw is additionally required. 
In our implementation, we first recover the relative rotation via essential matrix estimation. 
When the number of inliers is insufficient or the geometric estimation is unreliable, we fall back to a simple online calibration based on the observed direction of the current subgoal, from which the yaw is approximated. 
We note that this is a lightweight implementation, and more robust yaw estimation (e.g., learning-based approaches) can be incorporated, which is beyond the scope of this work.

Results are reported in Tab.~\ref{tab:learned_perception}. 
Under GT-PR, our method consistently outperforms all baselines across all tasks, although the margin is smaller compared to the GT-topological setting due to perception noise, highlighting the importance of robust perception. 
When replacing GT-PR with NetVLAD-based PR, performance generally degrades in most cases. 
However, our method with NetVLAD still outperforms all baselines that use GT-PR, demonstrating the effectiveness of intent-based guidance in improving policy robustness under imperfect perception.

Interestingly, in the \textit{Reverse} task, NetVLAD-based PR achieves better performance than GT-PR. 
This is expected since in the reverse setting, the agent’s viewpoint is largely opposite to the mapped trajectory. 
GT-PR retrieves the nearest map image based on position, which may not share visual overlap with the current observation, whereas NetVLAD retrieves visually similar images, providing more meaningful associations and improving navigation performance.

Overall, these results demonstrate that the proposed method remains effective under learned perception and shows strong potential for real-world deployment without reliance on simulator-provided ground-truth signals.


\section{Conclusion}

We introduce \emph{IntentReact}, an intent-conditioned object-centric navigation framework that bridges global topological guidance and local reactive control. 
By injecting a low-dimensional intent signal, the policy resolves directional ambiguity under partial observability while preserving reactivity.
Experiments show consistent gains across challenging settings, including large initial orientation errors and learned perception. 
The method remains robust under imperfect intent estimation, and ablations reveal that coarse directional guidance is sufficient, while BEV-based refinement further improves geometric consistency.
These results highlight that explicit directional intent provides a simple yet powerful mechanism to integrate global structure into reactive navigation, paving the way toward scalable and deployable navigation systems that operate reliably under real-world uncertainty. Future work includes improving perception and intent estimation under real-world noise and extending the framework to more complex and dynamic environments.


\bibliographystyle{IEEEtranBST/IEEEtran}
\bibliography{IEEEtranBST/IEEEabrv,IEEEtranBST/bare_jrnl_new_sample4} 

\end{document}